\documentclass{Odyssey2026}

\usepackage[table]{xcolor}
\usepackage{booktabs}
\usepackage{array}
\usepackage{amsfonts} 
\usepackage{amsmath, amssymb}
\definecolor{headerblue}{HTML}{003366}
\definecolor{rowgray}{HTML}{F2F2F2}
\usepackage{pdflscape}
\usepackage{tabularx}
\usepackage{booktabs}

\interspeechcameraready 

\title{End-to-End Intracortical Speech Decoding from Neural Activity}

\author[affiliation={1}]{Owais M.}{Khanday}
\author[affiliation={1}]{Jose A.} {Gonzalez-Lopez}
\author[affiliation={2}]{Marc} {Ouellet}
\author[affiliation={3}]{Alberto} {Galdón}
\author[affiliation={3}]{Gonzalo} {Olivares Granados}

\affiliation{Dpt. of Signal Theory, Telematics and Communications}{University of Granada}{Spain}
\affiliation{Brain, Mind, and Behavior Research Center}{University of Granada}{Spain}
\affiliation{Grupo CSUR de Epilepsia Refractaria}{Hospital Virgen de Las Nieves (Granada)}{Spain}
\email{\{owaismujtaba,joseangl,mouellet\}@ugr.es, \{alberto.galdon.sspa,gonzalo.olivares.sspa\}@juntadeandalucia.es}

\usepackage{comment}
\usepackage{soul}
\usepackage{booktabs} 

\begin{document}

\maketitle

\begin{abstract}
Current high-performing intracortical speech neuroprostheses achieve low word error rates but typically rely on external language models during inference, increasing memory, computation, and latency. In this work, we investigate whether meaningful character-level decoding is achievable without such models. We propose an end-to-end Conformer-based neural decoder trained directly on intracortical recordings from a participant with amyotrophic lateral sclerosis (ALS). Without any external language model, the system achieves a character error rate (CER) of 23.80\% on held-out validation data. Analysis shows that performance variability is driven by inter-session signal degradation, while dominant errors arise from incorrect word boundary segmentation. These results demonstrate that effective character-level decoding is possible in a fully end-to-end framework, providing a strong neural signal for downstream linguistic processing.
\end{abstract}

\section{Introduction}
    Neural speech prostheses  \cite{chang2024brain,
silva2024speech} represent one of the most ambitious frontiers in modern neuroscience and biomedical engineering, offering the prospect of restoring lost communication to individuals with severe neurological conditions~\cite{hochberg2006neuronal, shenoy2013cortical, andersen2004intention,
    donoghue2002connecting}. Among the populations who stand to benefit most are those affected by amyotrophic lateral sclerosis (ALS), spinal cord injury (SCI), locked-in syndrome (LIS), or brainstem stroke, conditions that progressively or abruptly sever the motor pathways through which humans express intent, yet leave higher cognitive function largely intact~\cite{birbaumer2000slow, chaudhary2016locked, vansteensel2016fully}. For these individuals, the inability to speak, write, or gesture does not reflect an absence of thought but rather an inability to translate intent into physical action. Speech neuroprostheses aim to bridge this gap by decoding neural signals directly from the brain, bypassing damaged motor pathways entirely~\cite{wolpaw2002brain, lebedev2006brain}. The field has explored a wide range of recording modalities, each offering a different trade-off between signal resolution, invasiveness, and long-term stability~\cite{leuthardt2004brain}. Non-invasive approaches such as electroencephalography (EEG) and functional magnetic resonance imaging (fMRI) offer accessibility and safety but are limited by low spatial resolution and susceptibility to noise~\cite{wang2022open, tang2023semantic, nakanishi2018enhancing, abiri2019comprehensive}.

    Electrocorticography (ECoG), which involves placing electrode grids directly on the cortical surface, achieves substantially higher signal fidelity and has enabled impressive speech and language decoding results~\cite{makin2020machine, moses2021neuroprosthesis, herff2015brain, angrick2019speech}. At the highest end of the  resolution spectrum, intracortical microelectrode arrays record single-unit and
    multi-unit spiking activity from small populations of neurons with millisecond temporal precision, providing the richest neural signal available to date~\cite{pandarinath2017high, gilja2015clinical, hochberg2012reach, trautmann2019accurate}. 
    
    A series of landmark studies in recent years has helped define the current state of the art in intracortical speech neuroprostheses~\cite{willett2021high, willett2023high, Card2024}. Willett et al.~\cite{willett2021high} demonstrated that neural activity recorded from the hand-knob area of the motor cortex could be decoded into handwritten characters at approximately 90 characters per minute with high accuracy, matching the speed of able-bodied typists and establishing the motor cortex as a rich source of decodable linguistic intent. Subsequently, Willett et al.~\cite{willett2023high} extended this paradigm to direct speech decoding, achieving real-time synthesis of attempted speech from a participant with ALS at 62 words per minute with a word error rate of 23.8\% on a large vocabulary, a performance level approaching that of contemporary automatic speech recognition (ASR) systems~\cite{baevski2020wav2vec, radford2023robust}. Card et al.~\cite{Card2024} further advanced this line of work, reporting sub-5\% word error rates sustained over 248 hours of naturalistic conversation, demonstrating that intracortical speech BCIs can achieve clinically meaningful accuracy over extended deployment periods.

    Despite these impressive achievements, a common architectural characteristic of current state-of-the-art intracortical speech decoding systems is their reliance on external language models (LMs) to rescore the output of a connectionist temporal classification (CTC)~\cite{graves2006connectionist} beam search at inference time. In the system of Willett et al.~\cite{willett2023high}, a Transformer-based language model with a vocabulary of 125,000 words is used to re-rank candidate transcriptions, contributing substantially to the final word error rate reduction. Similarly, Card et al.~\cite{Card2024} employ an $n$-gram language model integrated into the decoding pipeline.

    While external language models are highly effective for improving word-level accuracy, they introduce substantial practical constraints. Large-vocabulary language models require significant memory, and CTC beam search with LM integration is inherently sequential and not easily parallelizable, introducing latency proportional to beam width~\cite{hannun2014deep}. For fully implanted speech neuroprostheses, where decoding hardware must be miniaturized, power-efficient, and capable of real-time operation without a tethered external computer, these requirements may become prohibitive~\cite{shenoy2013cortical, kao2020considerations}. Importantly, the role of the LM in these systems is to impose linguistic structure and correct decoding errors, rather than to extract information directly from the neural signal. This raises a key question: to what extent can neural decoding performance be improved in an end-to-end framework without relying on external LM rescoring?

    The central motivation of this work is therefore to investigate whether meaningful character-level intracortical speech decoding is achievable in a fully end-to-end setting. To this end, we propose a Conformer-based~\cite{gulati2020conformer} sequence decoder that maps intracortical neural recordings directly to character sequences using a CTC objective~\cite{graves2006connectionist}, without any LM at inference time. Furthermore, in order to address the well-documented challenge of inter-session neural non-stationarity, arising from electrode drift, impedance fluctuations, and day-to-day changes in neural population dynamics~\cite{simeral2011neural, gallego2017neural, chestek2011single}, we introduce a session-specific linear alignment layer that precedes the shared Conformer encoder. This lightweight adapter allows the model to normalize session-specific feature distributions while the shared encoder learns more stable, session-invariant representations~\cite{li2019jasper}. We further develop a targeted data augmentation strategy to improve generalization across 45 recording sessions spanning 20 months of clinical data~\cite{park2019specaugment}. Our results demonstrate that meaningful character-level decoding is achievable without external LM rescoring, providing a promising step towards simpler and more deployable intracortical speech neuroprostheses.

    The rest of the paper is organized as follows. Section~\ref{sec:related-work} reviews prior work in neural speech decoding. Section~\ref{sec:methods} presents the proposed methodology, including the participant and dataset, model architecture, and data augmentation strategy. Section~\ref{sec:results} reports the experimental results, and Section~\ref{sec:conclusion} concludes the paper with limitations and future directions.

\section{Related Work}
\label{sec:related-work}

The decoding of speech and language from neural signals has progressed rapidly across multiple recording modalities~\cite{brumberg2010brain, martin2014decoding, anumanchipalli2019speech}. Early work with ECoG demonstrated that neural activity in speech-related cortical regions contains sufficient information to reconstruct acoustic features and classify phonemes~\cite{herff2015brain, anumanchipalli2019speech}. Sequence-to-sequence approaches further enabled direct neural-to-text mapping in constrained vocabularies, although limited lexical coverage restricted their applicability to open-ended communication~\cite{makin2020machine}.

Subsequent studies focused on clinically relevant scenarios. Moses et al.~\cite{moses2021neuroprosthesis} demonstrated real-time communication in a participant with anarthria using a CNN--LSTM word classifier over a small vocabulary, while later work increased communication rates and vocabulary size~\cite{metzger2023high}. 

In the intracortical domain, recent advances have achieved substantial improvements in decoding accuracy and speed. Willett et al.~\cite{willett2021high} showed high-rate character-level decoding from motor cortex activity, and later extended this approach to direct speech decoding using CTC-based models combined with language model rescoring~\cite{willett2023high}. Card et al.~\cite{Card2024} further improved performance using Conformer-based encoders, achieving low word error rates over long-term recordings.

A common architectural paradigm has emerged in these systems: a neural encoder produces phoneme or character probabilities, which are combined with a language model during decoding~\cite{hannun2014deep, chan2016listen}. This approach, inherited from automatic speech recognition (ASR), leverages the statistical structure of language to improve word-level accuracy.

From a systems perspective, however, integrating language models introduces additional computational and memory requirements, as well as decoding latency due to beam search~\cite{hannun2014deep}. While effective, this coupling motivates the exploration of approaches that strengthen the neural decoding component itself, enabling a clearer separation between neural signal extraction and downstream linguistic processing.

Another key challenge in intracortical speech neuroprostheses is the non-stationarity of neural signals across recording sessions~\cite{simeral2011neural, gallego2017neural}. Variability arises from electrode drift, changes in neural firing patterns, and other biological factors, often degrading cross-session generalization. Prior work has addressed this through recalibration~\cite{jarosiewicz2015virtual}, manifold alignment~\cite{degenhart2020stabilization}, and lightweight adaptation layers~\cite{Card2024, li2019jasper}.

Finally, Conformer  architectures~\cite{gulati2020conformer} have recently emerged as effective encoders for neural-based speech decoding, combining self-attention and convolutional modules to capture both global and local temporal structure. Patch-based tokenization strategies further improve efficiency by reducing sequence length while preserving temporal context~\cite{dosovitskiy2020vit, liu2021swin}. In this work, we build on these ideas by combining a Conformer encoder with a session-specific alignment layer and temporal patch embedding for end-to-end character-level decoding.

\section{Methods}
\label{sec:methods}
\begin{figure}[t]
    \centering
    \includegraphics[width=0.5\textwidth]{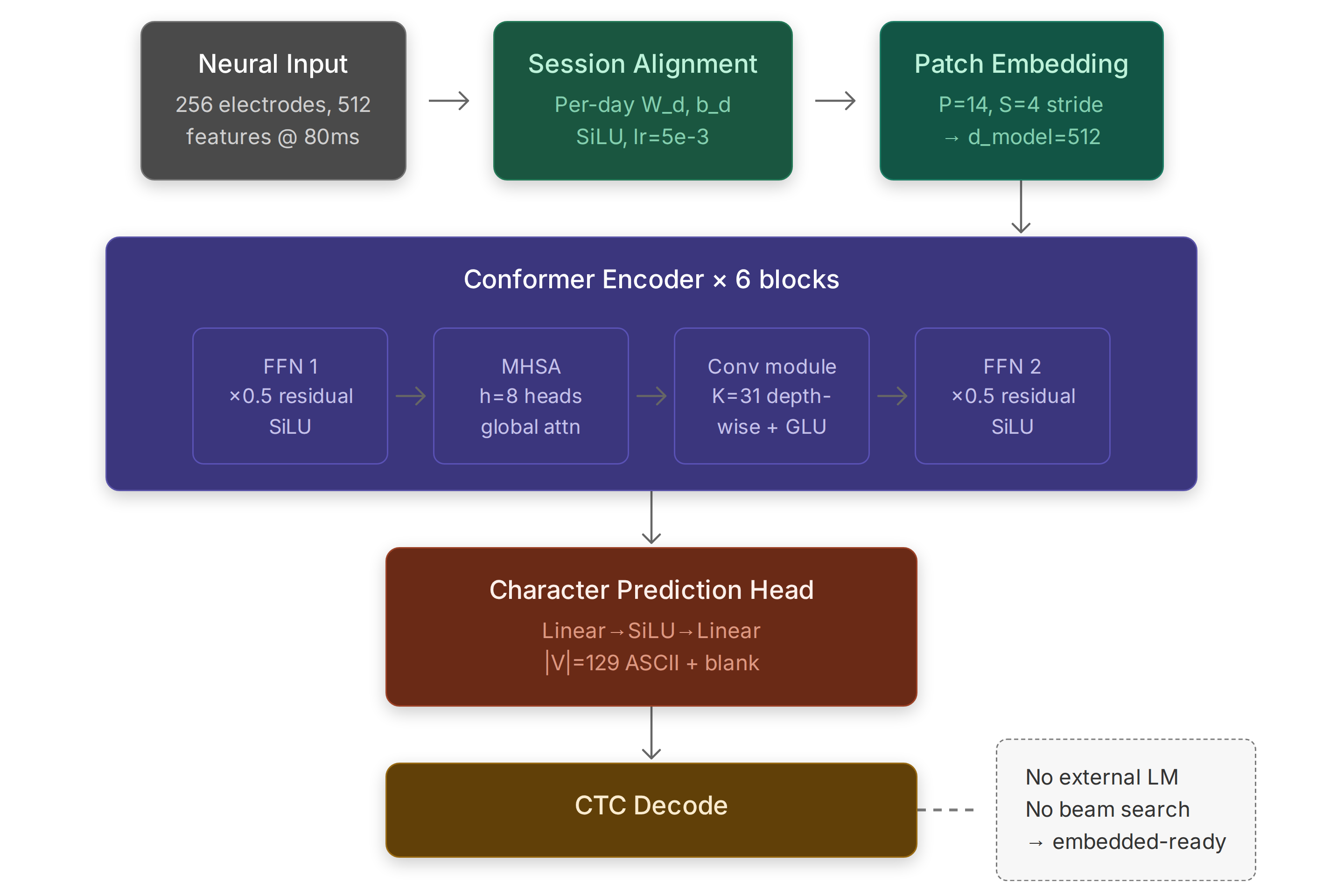}
    \caption{Overview of the proposed Conformer-based intracortical speech decoding architecture.}
    \label{fig:archi}
\end{figure}
    
We evaluate an end-to-end intracortical speech decoder on the public Brain-to-Text '25 benchmark. The proposed pipeline, depicted in Fig. \ref{fig:archi}, first applies a session-specific alignment layer to the neural features, followed by temporal patch embedding and a Conformer encoder that predicts character sequences with a CTC objective. During training, we apply a set of neural data augmentation strategies designed to improve robustness to inter-session variability and noise.

\subsection{Participant and Dataset}
\label{sec:dataset}
   
    In this study, we use the publicly available Brain-to-Text '25 benchmark\footnote{\url{https://www.kaggle.com/competitions/brain-to-text-25}}, a large-scale dataset for intracortical speech decoding with standardized training, validation, and test splits. The dataset consists of neural recordings and behavioral data from a 45-year-old male participant with ALS, as described in Card et al.~\cite{Card2024}, and spans 45 recording sessions collected over approximately 20 months. The participant exhibited severe dysarthria and tetraparesis but retained intact cognitive function.

    Neural activity was recorded using four 64-channel intracortical microelectrode arrays (256 electrodes in total) implanted in the left ventral precentral gyrus. These arrays target cortical regions involved in speech production, including the primary motor cortex (Brodmann area 4), ventral premotor cortex, and adjacent language-related areas.

    The recorded neural signals are represented using 512 features, comprising threshold crossing counts and spike band power extracted from each electrode. These features are computed in non-overlapping temporal bins of 20 ms, providing a high temporal resolution representation of population-level neural activity. This representation captures both spiking activity and local field potential dynamics relevant for speech-related motor planning and execution.

    The dataset contains a total of 10,948 trials, divided into 8,072 training trials, 1,426 validation trials, and 1,450 test trials. Each trial corresponds to a prompted speech utterance, with aligned text transcripts used as supervision for model training and evaluation.
    
    The implementation details can be found at https://github.com/owaismujtaba/E2ESpeechDecoding.git

\subsection{Model Architecture}
\label{sec:model}  
    We introduce a Conformer-based sequence decoder for an uninterrupted brain-to-text translation. The framework translates high-dimensional intracortical neural recordings to character-level transcriptions through a CTC objective~\cite{graves2006connectionist}. The model has four components, as depicted in Fig. \ref{fig:archi}: (i) a session-specific linear alignment layer, (ii) a strided temporal patch-embedding module, (iii) a stack of Conformer blocks, and (iv) a character prediction head. Model hyperparameters are  outlined in Table~\ref{tab:arch_summary}. In the following subsections, we describe these four components in detail.


    \newcolumntype{L}[1]{>{\raggedright\arraybackslash}p{#1}}

    \begin{table}[t]
        \centering
        \caption{Neural-decoding architecture configuration.}
        \label{tab:arch_summary}
        \small
        \begin{tabular}{L{3.2cm} L{4.3cm}}
        \toprule
        \textbf{Component} & \textbf{Configuration} \\
        \midrule
        Input Dimension   & 512 neural channels \\
        Session Alignment & $\mathbf{W}_d \in \mathbb{R}^{512\times512}$, $\mathbf{b}_d \in \mathbb{R}^{512}$, SiLU \\
        Patch Embedding   & $P=14$, $S=4$, lin. project. to 512 \\
        Positional Encoding & Sinusoidal (fixed), max length 5,000 \\
        Encoder Depth     & $N_L = 6$ Conformer blocks \\
        $d_{\text{model}}$ & 512 \\
        Attention Heads   & $h = 8$ \\
        $d_{\text{ff}}$    & 2,048 ($4\times$ expansion) \\
        Convolution Kernel & $K = 31$ (depthwise) \\
        Prediction Head   & Linear $\rightarrow$ SiLU $\rightarrow$ Dropout $\rightarrow$ Linear \\
        Vocabulary        & $|\mathcal{V}| = 129$ (ASCII + blank) \\
        Loss              & CTC + entropy regularization ($\lambda = 0.05$) \\
        Parameters        & $\approx 47$M \\
        \bottomrule
        \end{tabular}
    \end{table}
            
    \subsubsection{Session-Specific Input Alignment}
        Significant inter-session variability emerges out from electrode drift, impedance fluctuations, and day-to-day changes in neural signal statistics in the intracortical neural recordings. In order to fend off these non-stationarities, we carry out  a session-specific linear adapter layer \emph{prior} to the shared Conformer encoder. For a trial belonging to session index $d \in \{1, \dots, N_{\text{days}}\}$, the input features $\mathbf{X} \in \mathbb{R}^{T \times C}$ are transformed as:
        \begin{equation}
          \mathbf{X}' = \sigma\left(\mathbf{X} \mathbf{W}_d + \mathbf{b}_d\right),
          \label{eq:day_layer}
        \end{equation}
        where $C$ is the number of neural channels , $\mathbf{W}_d \in \mathbb{R}^{C \times C}$ is a session-specific weight matrix, $\mathbf{b}_d \in \mathbb{R}^{1 \times C}$ is a bias vector, and $\sigma(\cdot)$ denotes the SiLU (Sigmoid Linear Unit) activation function. 
        
        In order to enable stable convergence and maintain a "pass-through" state at the initial stage of training, we start each $\mathbf{W}_d$ as an identity matrix $\mathbf{I}$ and each $\mathbf{b}_d$ to zero. Throughout the forward pass duration, these parameters are dynamically indexed and applied via a clustered \texttt{einsum} process. 

        Optimization of these session-specific parameters is done independently from the shared network weights. The set $\{\mathbf{W}_d, \mathbf{b}_d\}$ is given to a devoted performance optimizer group within the AdamW optimizer. This group uses a higher autonomous learning rate ($\eta_{\mathrm{day}} = 5 \times 10^{-3}$) and a specific weight decay schedule ($\lambda_{\mathrm{day}}$). It permits the model to quickly align session-specific parameters.
    
        This group utilizes a higher independent learning rate ($\eta_{\mathrm{day}}$) and a specific weight decay schedule ($\lambda_{\mathrm{day}}$), allowing the model to rapidly align session-specific feature distributions while the shared encoder learns more stable, session-invariant representations.

    \subsubsection{Temporal Patch Embedding}
        Inspired by the patch-based tokenization in Vision Transformers~\cite{dosovitskiy2020vit}, we implement a strided temporal patch embedding to reduce the sequence length before it enters the Conformer encoder. Given the adapted neural features $\mathbf{X} \in \mathbb{R}^{T \times C}$ from the session-specific layer, we apply a 1-D unfolding operation along the temporal dimension with patch size $P$ and stride $S$. This operation yields a sequence of $N$ overlapping patches, where $N = \lfloor(T - P)/S\rfloor + 1$. Each patch $\mathbf{x}_n \in \mathbb{R}^{P \times C}$ is flattened into a vector $\mathbf{v}_n \in \mathbb{R}^{PC}$ and linearly projected into the model's latent space:
        \begin{equation}
          \mathbf{z}_n = \mathbf{v}_n \mathbf{W}_{\text{proj}} + \mathbf{b}_{\text{proj}},
          \label{eq:patch_embed}
        \end{equation}
        where $\mathbf{W}_{\text{proj}} \in \mathbb{R}^{PC \times d_{\text{model}}}$ and $\mathbf{b}_{\text{proj}} \in \mathbb{R}^{d_{\text{model}}}$ constitute the input projection layer. This design compresses the temporal resolution by a factor approximately equal to $S$, significantly reducing the computational complexity of the subsequent self-attention layers while allowing each token to capture local temporal dynamics. 
        
        Following the projection, a sinusoidal positional encoding $\mathbf{E}_{\text{pos}} \in \mathbb{R}^{N \times d_{\text{model}}}$ is element-wise added to the sequence to provide the model with relative temporal information:
        \begin{equation}
          \mathbf{Z}_{\text{input}} = \text{Dropout}(\mathbf{Z} + \mathbf{E}_{\text{pos}}),
        \end{equation}
        where $\mathbf{Z} = [\mathbf{z}_1, \dots, \mathbf{z}_N]^\top$. An input dropout layer is applied to the resulting embeddings to improve generalization before they are processed by the $L$ stacked Conformer blocks.

    \subsubsection{Conformer Encoder}
        The core encoder consists of $N_L = 6$ stacked Conformer blocks~\cite{gulati2020conformer}, implementing a \emph{Macaron-style} sandwich architecture. Each block processes the input sequence through four distinct sub-modules, using pre-normalization (LayerNorm) and residual connections:
         
        \begin{enumerate}[leftmargin=*, label=\arabic*.]
          \item \textbf{Feed-Forward Module 1.} 
            A half-step residual module consisting of a two-layer linear network with an expansion factor of 4. It applies $\text{Linear}(d_{\text{model}}, 4d_{\text{model}}) \to \text{SiLU} \to \text{Dropout} \to \text{Linear}(4d_{\text{model}}, d_{\text{model}}) \to \text{Dropout}$. The output is scaled by $0.5$ before the residual addition.
          
          \item \textbf{Multi-Head Self-Attention (MHSA).} 
            Standard multi-head attention with $h = 8$ heads and $d_{\text{model}} = 512$. This module captures global temporal dependencies across the entire patched sequence:
            \begin{equation}
              \mathbf{x} \leftarrow \mathbf{x} + \text{Attention}(\text{LN}(\mathbf{x}), \text{LN}(\mathbf{x}), \text{LN}(\mathbf{x})).
            \end{equation}
         
          \item \textbf{Convolutional Module.} 
            A sequence-to-sequence transformation designed to capture local articulatory patterns. It utilizes a point-wise $1{\times}1$ convolution to expand channels to $2d_{\text{model}}$, followed by a Gated Linear Unit (GLU) activation, which reduces the dimension back to $d_{\text{model}}$. A depthwise 1D convolution with kernel size $K = 31$ $groups=d_{\text{model}}$ is then applied, followed by Batch Normalization, SiLU activation, and a final $1{\times}1$ point-wise projection and dropout.
         
          \item \textbf{Feed-Forward Module 2.} 
            A second half-step FFN, structurally identical to Module 1, completing the Macaron-style block. A final Layer Normalization is applied at the output of the block.
        \end{enumerate}
         
        The total model capacity is defined by $d_{\text{model}} = 512$ and $d_{\text{ff}} = 2048$. Dropout with $p = 0.1$ (or as specified by hyperparameters) is applied within all sub-modules to prevent overfitting to session-specific noise.

    \subsubsection{Character Prediction Head}
        The output of the final Conformer block is passed through a two-layer prediction head:
        \begin{equation}
            \mathbf{y} = \mathbf{W}_2 \, \operatorname{SiLU} \left( \mathbf{W}_1 \mathbf{h} + \mathbf{b}_1 \right) + \mathbf{b}_2,
            \label{eq:head}
        \end{equation}
        where $\mathbf{W}_1 \in \mathbb{R}^{d_{\text{model}} \times d_{\text{model}}}$ and 
        $\mathbf{W}_2 \in \mathbb{R}^{d_{\text{model}} \times |\mathcal{V}|}$.
        The vocabulary encodes $|\mathcal{V}| = 129$ tokens: the 128 printable ASCII characters shifted by 32 $+1$ (so that raw ASCII value $c$ maps to index $c + 1$), with index~0 reserved as the CTC blank token. Log-softmax is applied to produce log-probabilities for CTC decoding.

    \subsubsection{Training Objective}
        The model is trained end-to-end with the CTC loss originally proposed in~\cite{graves2006connectionist} using blank index~0 and zero-infinity handling. To discourage the model from collapsing to over-peaked blank predictions, an entropy-regularization term is added:
        \begin{equation}
          \mathcal{L} = \mathcal{L}_{\mathrm{CTC}}
          + \lambda \cdot \bigl(-\mathbb{E}[\log p(\mathbf{y}\mid\mathbf{x})]\bigr),
          \label{eq:loss}
        \end{equation}
        where $\lambda = 0.05$ is the label-smoothing coefficient. This term  penalizes over-peaked output distributions and acts as output-entropy  regularization, analogous to label smoothing in classification tasks.
        A summary of the architecture configuration is shown in Table \ref{tab:arch_summary}

\subsection{Data Augmentation}
\label{sec:data-augment}
    Intracortical arrays provide neural recordings that are subject to considerable trial-to-trial and session-to-session variability. These include additive noise from thermal and biological processes, incremental drifts in baseline firing rates, electrode impedance fluctuations, and trial-to-trial differences in movement kinematics. In order to enhance the resilience and extrapolation of the decoder, we employ a set of data augmentation methods to the neural attributes at the time of training. The whole set of augmentations is applied in the GPU feature space following loading but before the forward pass. At the time of validation and inference, only Gaussian smoothing is used.

    \subsubsection{Additive Noise Augmentations}
        Three forms of additive noise are applied independently to the feature tensor $x \in \mathbb{R}^{B \times T \times C}$:

        \begin{itemize}
            \item \textbf{White Noise:} Independent Gaussian noise $\epsilon \sim \mathcal{N}(0, \sigma^2_w)$ is added to every element of the feature tensor. This simulates channel-level recording noise and prevents the model from overfitting to precise spike amplitude values.
            \item \textbf{Constant Offset Noise:} A per-channel, per-trial offset $\delta \sim \mathcal{N}(0, \sigma^2_c)$ is sampled once per batch element and broadcast across all time steps. This simulates slowly drifting electrode baselines.
            \item \textbf{Random Walk Noise:} A cumulative sum of Gaussian increments along the time axis simulates non-stationary drift. 
        \end{itemize}

    \subsubsection{Temporal Augmentations}
        Three augmentations manipulate the temporal structure of each trial:
        
        \begin{itemize}
            \item \textbf{Random Temporal Cutoff:} A random number of time steps is removed from the beginning of each trial. This simulates variability in cue-to-movement onset latency and prevents the decoder from relying on absolute temporal position for early frames.
            \item \textbf{Time Warping:} Each trial is independently time-warped by a factor $w \sim \text{Uniform}(1 - \alpha, 1 + \alpha)$ using linear interpolation. The warped sequence is either zero-padded or truncated to the maximum buffer length. This augmentation encourages invariance to minor speed variations in attempted speech production.
            \item \textbf{Time Masking (SpecAugment-style):} Following the SpecAugment framework \cite{park2019specaugment} adapted for neural data, $N_m$ time masks are applied per trial. Each mask zeroes out a contiguous block of up to $L_m = 20$ consecutive time steps at a randomly chosen start position. This forces the model to rely on distributed temporal evidence rather than short, highly informative windows.
        \end{itemize}

    \subsubsection{Channel Augmentations}
        A channel dropout mask is applied by independently zeroing each channel with probability $p_{drop}$. The mask is broadcasted across time, simulating electrode failures or signal loss on individual channels during recording. This regularization promotes the learning of redundant, distributed representations across the electrode array.

        \subsubsection{Gaussian Smoothing}
        After all stochastic augmentations, a fixed Gaussian kernel is convolved along the time axis of the feature tensor to smooth the binned spike rates. The kernel has standard deviation $\sigma_s $ time steps and a support of $K = 100$ points, applied as a depthwise convolution with same padding. Smoothing is applied in both training and evaluation modes. This operation reduces high-frequency noise in the spike rate estimates and provides a smoother temporal trajectory for the encoder to process.

    The integration of these augmentation strategies is critical for addressing the inherent non-stationarity and low signal-to-noise ratio (SNR) of intracortical neural recordings. By injecting \textit{Additive Noise} and \textit{Constant Offsets}, the model is forced to marginalize over electrode-specific instabilities and biological background noise, preventing overfitting to transient signal artifacts. Furthermore, \textit{Temporal Augmentations} and \textit{Channel Dropout} ensure the decoder learns robust, distributed representations that remain invariant to fluctuations in movement timing and potential hardware failures. Ultimately, these techniques facilitate superior generalization from constrained training sets to real-time inference, where environmental and physiological conditions frequently deviate from the calibration phase. A snapshot of the augmentation values are presented in Table \ref{tab:augmentation_summary}

    \begin{table}[t]
        \centering
        \caption{Hyperparameter configurations for data augmentation. All stochastic augmentations are restricted to the training phase to improve model robustness, while Gaussian smoothing is applied across both training and validation to ensure stable firing rate estimates.}
        \label{tab:augmentation_summary}
        \begin{tabular}{@{}llll@{}}
        \toprule
        \textbf{Augmentation} & \textbf{Parameter} & \textbf{Value} & \textbf{Applied At} \\ \midrule
        White Noise & $\sigma_w$ & 1.0 & Train only \\
        Constant Offset & $\sigma_c$ & 0.2 & Train only \\
        Random Temporal Cut & max cut & 3 steps & Train only \\
        Time Warping & $\alpha$ & 0.10 & Train only \\
        Time Masking & $N_m / L_m$ & 2 / 20 & Train only \\
        Channel Dropout & $p_{drop}$ & 0.05 & Train only \\
        Gaussian Smoothing & $\sigma_s / K$ & 2 / 100 & Train + Val \\ \bottomrule
        \end{tabular}
    \end{table}

\subsection{Evaluation Protocol}
We use the standard train/validation/test splits of the Brain-to-Text '25 benchmark. Model selection is based on validation CER, with final results reported from the best checkpoint (optionally using SWA).

Decoding is performed using greedy CTC without beam search or external language model integration. Performance is measured using character error rate (CER), computed as the normalized Levenshtein distance between predicted and reference sequences. We do not report word error rate (WER), as no language model is used during decoding. CER is used to isolate the performance of the neural decoder.

\section{Results}
\label{sec:results}

In this section, we report the performance of the proposed model on the Brain-to-Text '25 benchmark, focusing on the validation set (1,426 sentences), and analyze the main factors influencing its behavior.

\subsection{Overall Performance and Model Comparison}
\begin{table}[t]
\centering
\caption{Comparison of character error rate (CER) on the validation set.}
\label{tab:main_results}
\scriptsize
\begin{tabular}{llc}
\toprule
\textbf{Model} & \textbf{Architecture} & \textbf{CER (\%)} \\
\midrule
Baseline (Brain-to-Text '25) & GRU + CTC & 30.00 \\
\midrule
Conformer  & Conformer + align. + aug. & \textbf{23.80} \\
Conformer w/o alignment. & Conformer + aug. & \textbf{24.28} \\
Conformer w/o augmentation & Conformer + align. & \textbf{39.55} \\
\bottomrule
\end{tabular}
\end{table}

We first compare in Table \ref{tab:main_results} the proposed Conformer-based model against the baseline provided with the Brain-to-text'25 benchmark \cite{Card2024}, as well as ablated variants of our architecture. The proposed model achieves a minimum validation CER of \textbf{23.80\%}, outperforming the baseline system.

In particular, our full model yields an absolute reduction in CER of 23.80\% (relative improvement of 20.67\%) compared to the baseline, demonstrating the effectiveness of the proposed architecture. Training without data augmentation results in a performance drop of 15.75\% in CER, confirming that augmentation plays a key role in improving robustness and generalization.

Given that the full Conformer model achieves the best performance, the following analyses focus exclusively on this configuration.

\subsection{Session-wise Performance Variability}
\begin{figure}[t]
    \centering
    \includegraphics[width=0.5\textwidth]{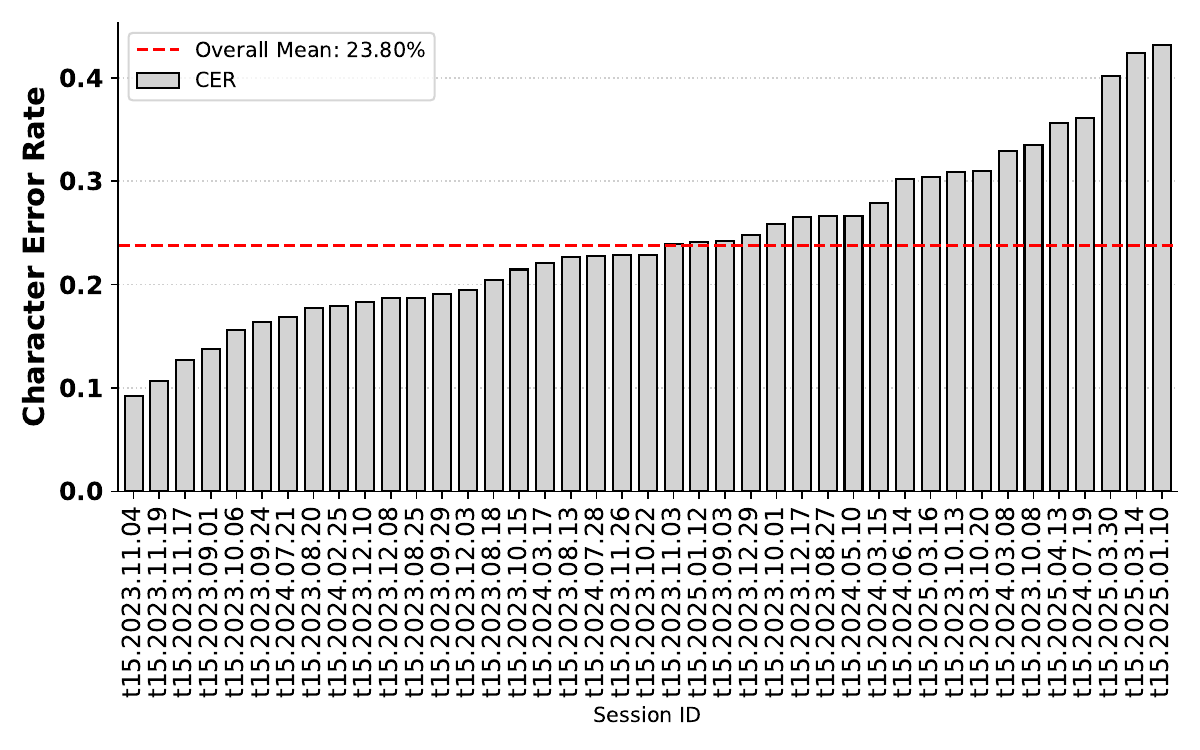}
    \caption{Mean CER per recording session.}
    \label{fig:sess_cer}
\end{figure}

We next analyze the robustness of the model across recording sessions. Fig.~\ref{fig:sess_cer} shows noticeable variability in CER across sessions, with a clear temporal trend: earlier sessions achieve CER values around 10\%, while later sessions reach 40--45\%. This pattern suggests a progressive degradation in signal quality over time.

Possible contributing factors include long-term electrode drift, reduced signal-to-noise ratio due to tissue encapsulation around the electrodes, and changes in the neural population being recorded. These effects may lead to less stable or less informative neural features, making decoding more challenging in later sessions.

\subsection{Utterance-level Error Distribution}

\begin{figure}[t]
    \centering
    \includegraphics[width=0.5\textwidth]{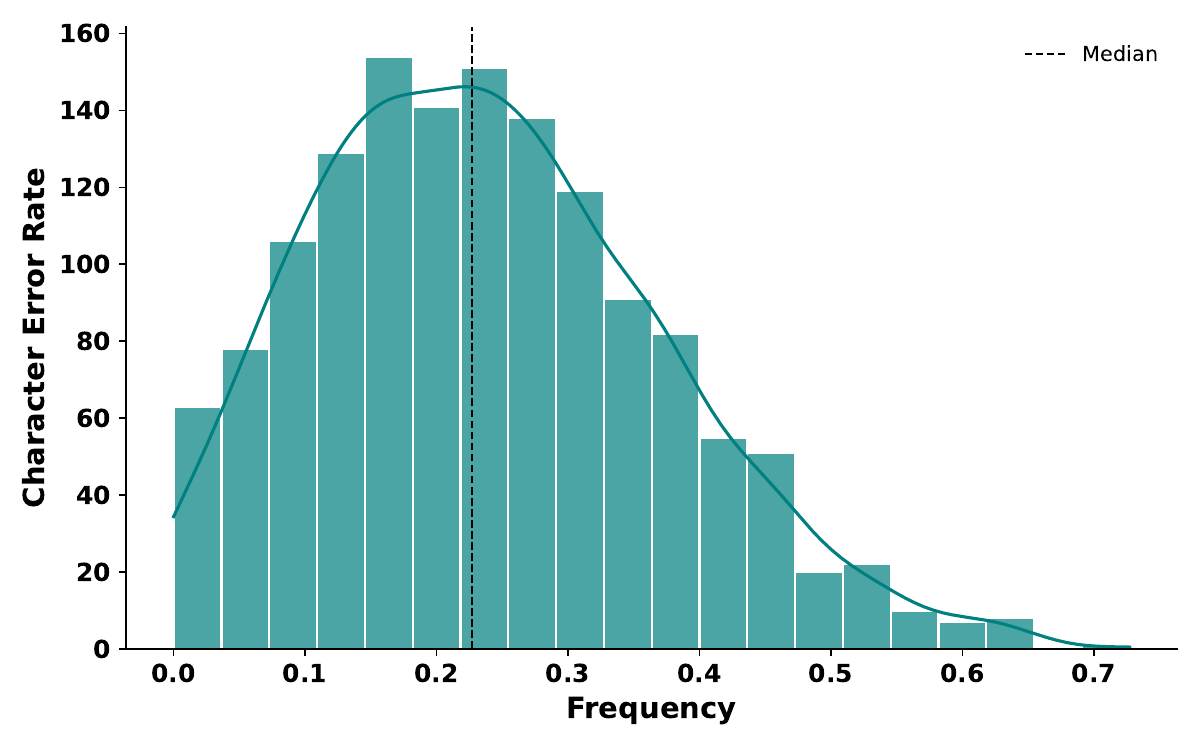}
    \caption{Distribution of CER across validation utterances.}
    \label{fig:sess_frew}
\end{figure}

Fig.~\ref{fig:sess_frew} presents the distribution of CER across utterances. The distribution is right-skewed, with most samples concentrated at low error rates and a long tail of high-error cases. The median CER is 0.22, with an interquartile range of 0.18.

\begin{figure}[t]
    \centering
    \includegraphics[width=\linewidth]{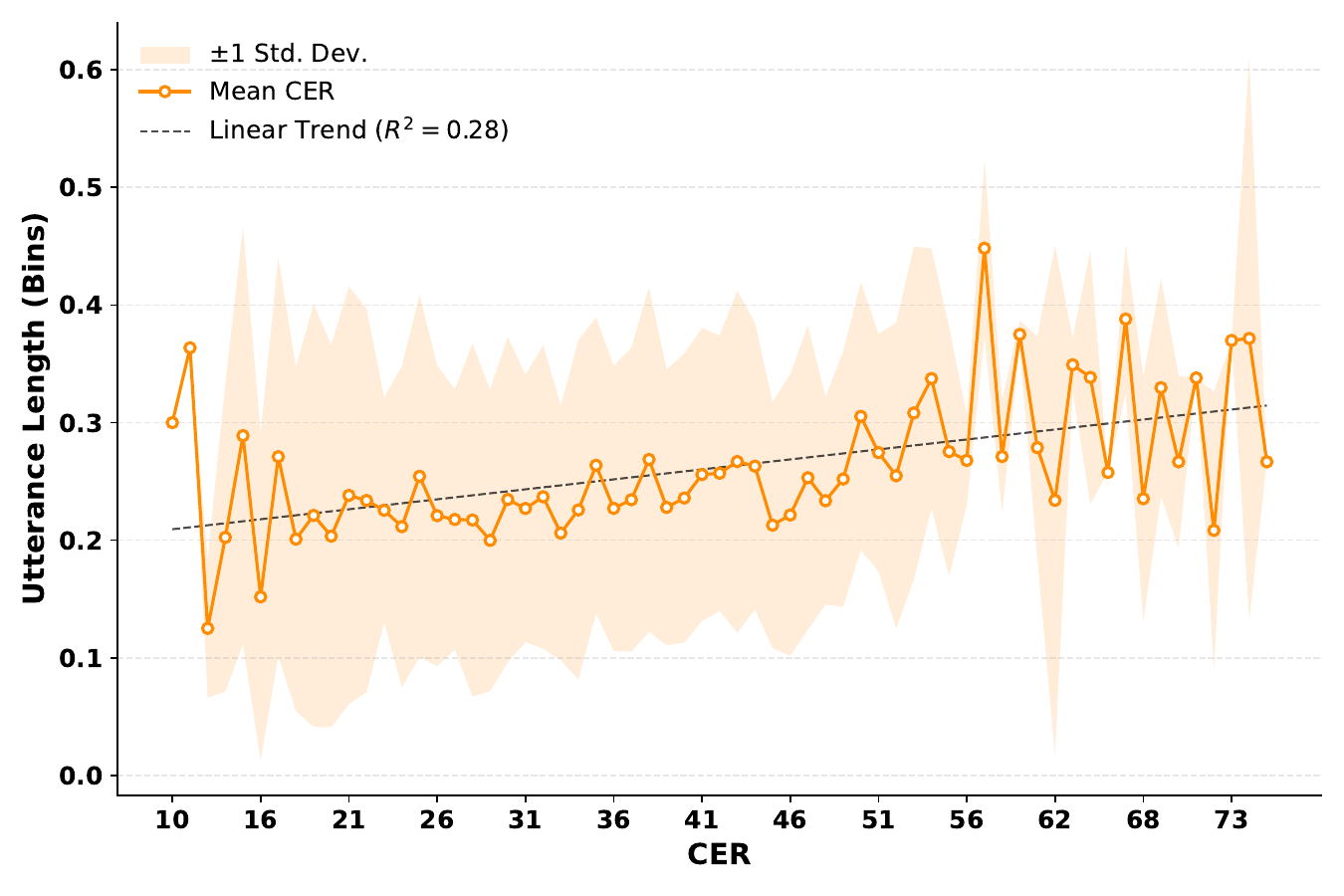}
    \caption{Mean CER as a function of utterance length (in characters). The shaded area represents $\pm$ one standard deviation, and the dashed line indicates a linear fit.}
    \label{fig:cer_length}
\end{figure}

Fig.~\ref{fig:cer_length} shows a weak relationship between utterance length and CER ($R^2 = 0.28$). Performance is more stable for mid-length utterances (approximately 20–50 characters), while short sequences exhibit higher variability due to the impact of single-character errors. For longer utterances, a gradual increase in CER is observed, suggesting that error accumulation over time reduces decoding reliability.

\subsection{Qualitative Analysis of Predictions}

\begin{table}[htbp] 
    \centering
    \scriptsize
    \caption{Representative qualitative examples (Validation Set). Incorrect words are highlighted in red.}
    \label{tab:qualitative_single}
    
    \begin{tabularx}{\columnwidth}{@{}XXcl@{}}
    \toprule
    \textbf{Ref.} & \textbf{Pred.} & \textbf{CER} & \textbf{Cat.} \\ \midrule
    
    How much is your card worth? & How much is your card worth? & 0.00 & Cor. \\ \addlinespace
    We think it's good. & We think it's good. & 0.00 & Cor. \\ \midrule
    
    I like to enjoy my life in the country. & I like to \textcolor{red}{enjoe} my \textcolor{red}{live} in the \textcolor{red}{cuntrry}. & 0.10 & Par. \\ \addlinespace
    She used to love to do stuff like that. & She \textcolor{red}{usd} to love to do \textcolor{red}{tuffe} \textcolor{red}{ligke} that. & 0.10 & Par. \\ \midrule
    
    Some new types of furniture. & \textcolor{red}{Thon ne lases are pier}. & 0.64 & Fail. \\ \addlinespace
    Not too long ago. & \textcolor{red}{Dit tyo w or?} & 0.65 & Fail. \\ \bottomrule
    
    \end{tabularx}
\end{table}

Table~\ref{tab:qualitative_single} presents representative examples of correct predictions, partial errors, and failure cases. Correct predictions typically correspond to phonetically clear and frequent expressions. Partial errors are dominated by local substitutions and deletions that preserve overall structure. In contrast, failure cases involve severe degradation and low overlap with the reference, often associated with short or uncommon utterances.

\subsection{Character-level Error Analysis}
\begin{table}[h]
\centering
\caption{Top 10 most frequent character-level substitution errors on the validation set.}
\label{tab:confusions}
\setlength{\tabcolsep}{8pt}
\renewcommand{\arraystretch}{1.2}
\begin{tabular}{clcr}
\toprule
\textbf{Rank} & \textbf{Ref $\rightarrow$ Hyp} & \textbf{Type} & \textbf{Count} \\
\midrule
1  & \texttt{e} $\rightarrow$ \texttt{[SP]}   & Deletion of boundary & 603 \\
2  & \texttt{[SP]} $\rightarrow$ \texttt{t}   & Insertion into word  & 500 \\
3  & \texttt{t} $\rightarrow$ \texttt{[SP]}   & Deletion of boundary & 476 \\
4  & \texttt{[SP]} $\rightarrow$ \texttt{e}   & Insertion into word  & 445 \\
5  & \texttt{[SP]} $\rightarrow$ \texttt{o}   & Insertion into word  & 442 \\
6  & \texttt{o} $\rightarrow$ \texttt{[SP]}   & Deletion of boundary & 378 \\
7  & \texttt{[SP]} $\rightarrow$ \texttt{a}   & Insertion into word  & 331 \\
8  & \texttt{s} $\rightarrow$ \texttt{[SP]}   & Deletion of boundary & 310 \\
9  & \texttt{n} $\rightarrow$ \texttt{[SP]}   & Deletion of boundary & 294 \\
10 & \texttt{[SP]} $\rightarrow$ \texttt{i}   & Insertion into word  & 278 \\
\midrule
\multicolumn{3}{l}{\textit{Total (top 10)}} & 4,057 \\
\bottomrule
\end{tabular}
\end{table}

    Finally, to better understand systematic decoding errors, we analyze character-level confusions. Table~\ref{tab:confusions} shows the most frequent substitution errors. Notably, all top errors involve the space token, indicating that the dominant failure mode is incorrect word boundary segmentation rather than phoneme-level confusion.

    The model frequently merges adjacent words (space deletion) or inserts spurious spaces within words, disrupting lexical structure. This pattern suggests that the acoustic–neural evidence for word boundaries is relatively weak compared to within-word character transitions. In particular, the confusion between spaces and high-frequency characters (e.g., \textit{e}, \textit{t}, \textit{o}, \textit{a}) indicates that boundary decisions are often replaced by more probable character emissions.
    
    This behavior is consistent with known limitations of CTC-based decoders, where alignment is learned implicitly and no explicit mechanism enforces consistent segmentation. As a result, errors tend to concentrate at word boundaries rather than within well-supported character sequences, which remain comparatively stable.

\section{Conclusion}
\label{sec:conclusion}

In this work, we presented an end-to-end Conformer-based decoder for intracortical speech neuroprostheses that directly maps neural activity to character sequences. By combining  dataset augmentation, a session-specific alignment layer, temporal patch embedding, and a Conformer encoder trained with a CTC objective and entropy regularization, the proposed model achieves a validation CER of 23.80\% on the Brain-to-text'25 benchmark. Ablation results further highlight the importance of both the alignment mechanism and the tailored data augmentation pipeline for achieving robust performance. Our analysis shows that performance varies significantly across recording sessions, with a clear degradation in later sessions, likely reflecting non-stationarity and potential declines in signal quality. At the character level, errors are dominated by incorrect word boundary segmentation, with frequent confusions between spaces and high-frequency characters, rather than purely phonetic substitutions.

Future work will focus on improving robustness to long-term signal drift and addressing segmentation errors, for instance by incorporating explicit word-boundary modeling or lightweight language constraints. Additional directions include scaling model capacity and exploring self-supervised pretraining on larger neural datasets. 

\section{Acknowledgement}

This work was supported by grants PID2022-141378OB-C22 and AIA2025-163317-C32 funded by MICIU/AEI/10.13039/501100011033 and ERDF/EU.

\bibliographystyle{IEEEtran}
\bibliography{Odyssey2026_BibEntries}

\end{document}